\documentclass[twocolumn]{article}
\usepackage{graphicx}

\title{A Real--Time Novelty Detector For A Mobile Robot \footnote{In Proceedings of the EUREL European Advanced Robotics Systems Masterclass and Conference, 2000}}
\author{Stephen Marsland, Ulrich Nehmzow and Jonathan Shapiro\\
Department of Computer Science\\University of Manchester\\Oxford Road\\Manchester M13 9PL, U.K.\\
\texttt{\{smarsland, ulrich, jls\}@cs.man.ac.uk}}
\date{}

\begin{document}
\maketitle

\thispagestyle{empty}

\begin{abstract}
{\em
Recognising new or unusual features of an environment is an ability 
which is potentially very useful to a robot. This paper demonstrates
an algorithm which achieves this task by learning an internal representation of
`normality' from sonar scans taken as a robot explores the environment.
This model of the environment is used to evaluate the novelty of each
sonar scan presented to it with relation to the model. 
Stimuli which have not been seen before, and therefore have more 
novelty, are highlighted by the filter. 
The filter has the ability to forget about features which have been
learned, so that stimuli which are seen only rarely
recover their response over time. A number of robot experiments
are presented which demonstrate the operation of the filter.

\vspace{5mm}
}
 
\end{abstract}

\section{\textbf{Introduction}}

Novelty detection, recognising when a particular stimulus has not been seen before,
is a very useful ability for both animals and robots. 
This paper presents
an algorithm which allows a robot to detect novel stimuli. 
The novelty filter described learns a representation of an
environment and then detects deviations from that model by evaluating the
novelty of each feature presented.

A novelty filter has many potential uses on a mobile robot. For instance, it could
be used as an attentional mechanism, directing the
robot's attention to newer features, which may be important and have not
previously been learned~\cite{Marsland00}. This reduces the amount of processing needed to
deal with the robot's sensory perceptions. The novelty filter could also 
enable the robot to be used as an inspection
agent. A model is built by the robot of a `clean' area, which has been inspected
by humans and is known to exhibit no undesirable features. The robot then 
explores the wider environment and marks those stimuli which are not present
in the model and therefore were not in the original environment.

This paper demonstrates the behaviour of the novelty filter when
the inputs to it are sonar scans taken while a robot explores an
environment using a wall-following behaviour. One property of the 
novelty filter which is investigated here is the ability to forget.
This means that it will still find to be novel any stimuli which are seen only infrequently.
This is useful because it ensures that these features are
always considered novel, not learned over time. This can help the robot to deal with dynamic
environments, where things may change over time. If an event happens
only occasionally we would like it to be considered novel, but without
forgetting the robot will learn to recognise it no matter what the
time interval between occurrences.

\subsection{\textbf{Related Work}}

The best known example of a novelty detector is the 
Kohonen Novelty Filter~\cite{Kohonen76,Kohonen93}.
This is an autoencoder neural network which is trained using backpropagation
of error~\cite{Bishop95b}. Once the network has been trained, presenting
an input to the network produces one of the learned outputs, and
taking the bitwise difference between the two displays the novel
components of the input.

A number of other researchers have proposed novelty filters. Ypma and Duin~\cite{Ypma97}
proposed a novelty detector based on the self-organising map. Training
data was used to train the map, so that the data formed organised 
neighbourhoods. Then, when any data caused a neuron to fire which was beyond a
predefined threshold from any of the neighbourhoods, the data was 
considered to be novel. This technique depends very strongly on the
choice of threshold and assumes that the data presented to the
network formed strictly segmented neighbourhood clusters. The technique
has been used by Taylor and MacIntyre~\cite{Taylor98} to detect machinery faults.
The network was trained on data recorded while the machine was operating
without problems, and data deviating from this pattern was taken as novel.

The technique of training the network on `normal' data and then attempting
to recognise whether inputs come from the learned probability distribution
is a common one when there is little data from a particular class, such as
machine faults, but lots of data from the other classes. It has been used
for topics as diverse as mammogram scans~\cite{Tarassenko95} to machine
breakdowns~\cite{Nairac99,Worden99}.

An alternative method was proposed by Ho and Rouat~\cite{Ho98} 
whose model is based on an integrate-and-fire
network. The algorithm times how long it takes the oscillatory
network to settle to a stable solution, reasoning that inputs
which have been seen previously will converge faster than
novel ones.

\section{\textbf{The Novelty Filter \label{HSOM}}}

The novelty filter described in this paper works on the principle that
something is novel if it has not been seen before. The question
is how to recognise that an item is new. If we know in advance what everything
in the environment will look like then it is relatively simple to train the
robot to recognise each of those features. However, this is usually not possible. 
Instead, if the robot learns to ignore anything which it has seen before,
then, it will only respond to novel things. 
This is something which animals do quite well~\cite{OKeefe77}.
There are then two parts to the desired system - learning to recognise features
that have been seen before, and evaluating their novelty. The first part, recognising
features, is a pattern recognition problem and has been considered
widely in the neural network literature. One possible solution is the
Kohonen Self-Organising Map~\cite{Kohonen93}, which is described below.
The second problem, how to evaluate the novelty, is considered in section~\ref{Eval}.

\subsection{The Self-Organising Map}

The Self-Organising Map (SOM) of Kohonen~\cite{Kohonen93} is a
clustering mechanism which clusters input vector in a topological
way, so that perceptions which are similar excite similar regions
of the network. The  SOM is used here to perform Learning Vector
Quantisation, choosing a winning neuron that best matches
the input and moving that neuron and its neighbours closer to
the input vector. It does this by selecting the neuron with 
the minimum distance between itself and the input. 
The distance is defined by:

\begin{equation}
d = \sum_{i=0}^{N-1} \left( \mathbf{w}_i (t) - \mathbf{v} (t) \right) ^2,
\label{dist}
\end{equation}

\noindent
where $\mathbf{v} (t)$ is the input vector at time $t$,
$\mathbf{w}_{i}$ the weight between input $i$ and the neuron and
the sum is over the $N$ components of the input vector.
The weights for the winning neuron and its eight topological neighbours
are updated by:

\begin{equation}
\mathbf{w}_{i} (t+1) = \mathbf{w}_{i} (t) + \eta (t) \left( \mathbf{v} (t) - \mathbf{w}_{i} (t) \right)
\end{equation}

\noindent
where $\eta$ is the learning rate, $0 \leq \eta (t) \leq 1$.
A square map field,
comprising 100 neurons arranged in a 10 by 10 grid, was used in
the experiments reported here. 
The neighbourhood size was kept constant at $\pm 1$ unit and the
learning rate $\eta$ was 0.25, so that the network was always learning.

\subsection{Evaluating the Novelty \label{Eval}}

\begin{figure}
\centering
\includegraphics[angle=270,width=.45\textwidth]{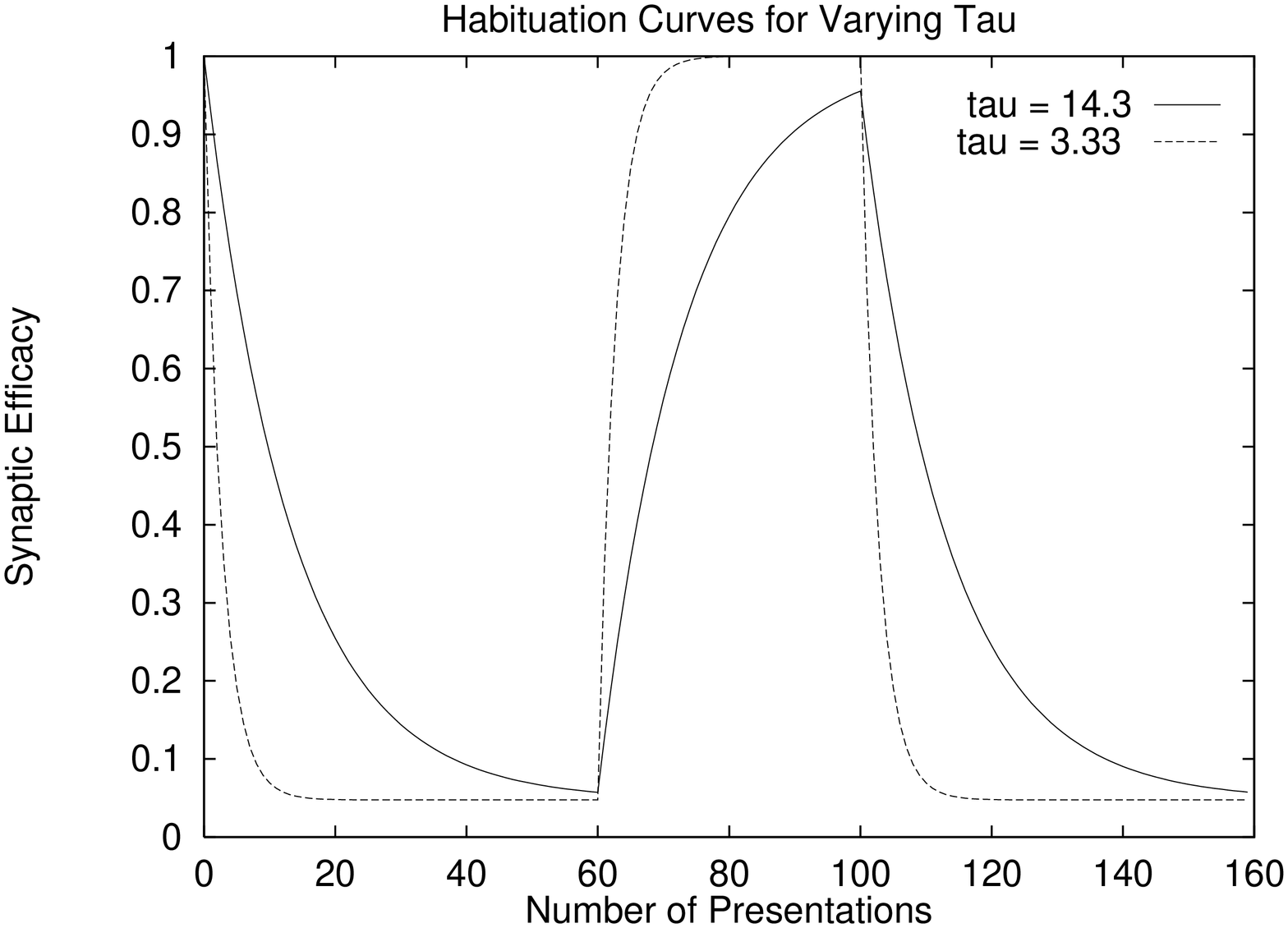}
\caption{
\textsf {\small {\em Left:} An example of how the synaptic efficacy drops when habituation occurs. 
In both curves, a constant stimulus $S(t)=1$ is presented,
causing the efficacy to fall. The stimulus is reduced to $S(t)=0$
at time $t=60$ where the graphs rise again, and becomes
$S(t)=1$ again at $t=100$, causing another drop. 
The two curves show the effects of varying $\tau$ in equation~\ref{HabEqn}.
It can be seen that a larger value of $\tau$ causes both the learning
and forgetting to occur faster. The other variables were the same for
both curves, $\alpha = 1.05$ and $y_0 = 1.0$.}}
\label{curves}
\end{figure}

Once a feature has been classified using the SOM, the novelty filter 
needs to assign a novelty value to the reading. A simple counter could
be kept on each neuron, recording the number of times that each neuron
has fired, and the output reduced accordingly. This is biologically implausible
and does not allow for any forgetting of stimuli.
When an animal stops responding to a feature which
has been presented to it repeatedly, the animal is said to have habituated to
the signal. Habituation, thought to be one of the simplest forms of plasticity
in the brain~\cite{Thompson66}, has been detected in a wide range of animals from the sea slug
{\em Aplysia}~\cite{Bailey83,Greenberg87} to humans~\cite{OKeefe77}.
The increase in the response to an habituated stimulus when the stimulus
is withdrawn is called dishabituation. It is thought to be a separate process
acting on the habituable synapses~\cite{Groves70}.

\begin{figure}
\centering
\includegraphics[angle=270,width=.4\textwidth]{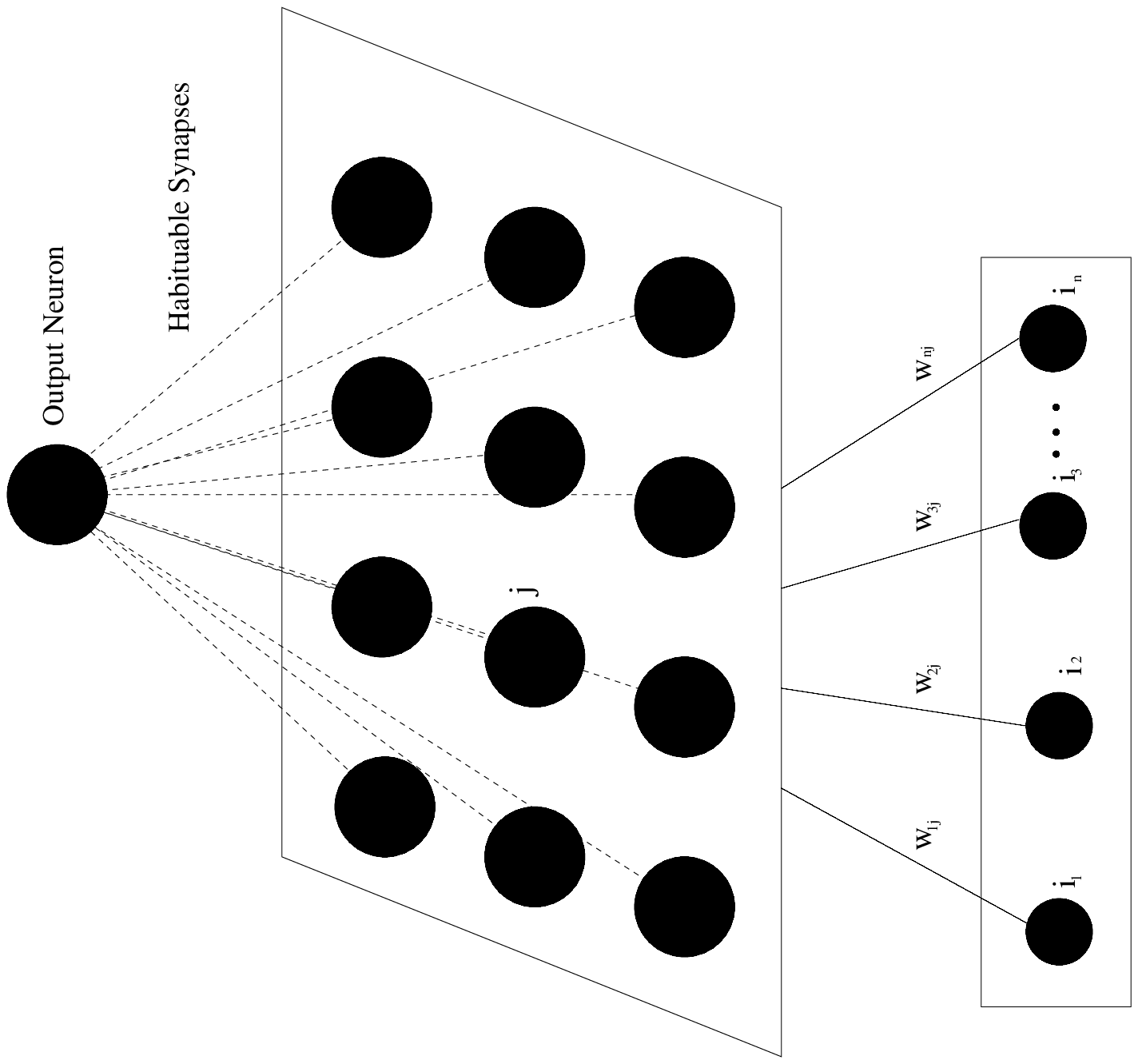}
\caption{ \textsf{ \small {The novelty filter.
The input layer connects to a clustering layer which
represents the feature space, the winning neuron (i.e.,
the one `closest' to the input) passing its
output along a habituable synapse to the output neuron so
that the output received from a neuron reduces with the
number of times it fires.}}}
\label{hsom}
\end{figure}

Several researchers have proposed models of the phenomenon of habituation, including
Groves~\cite{Groves70}, Wang and Hsu~\cite{Wang90} and Stanley~\cite{Stanley76}.
It is the model of Stanley, described below, which is used here. 
The synaptic efficacy, $y(t)$, decreases according to the following
equation:

\begin{equation}
 \tau \frac{dy(t)}{dt} = \alpha \left[ y_0 - y(t) \right] - S(t),
\label{HabEqn}
\end{equation}

\noindent
where $y_0$ is the original value of $y$, $\tau$ and $\alpha$
are time constants governing the rate of habituation and 
recovery respectively, and $S$ is the stimulus presented.
The activity of the winning neuron and its neighbours are propagated up the
synapse, so the input is $S(t) = d$ ($d$ defined in equation~\ref{dist}).
Using equation~\ref{HabEqn} we can control how strongly a synapse
responds to an input. The first time a synapse fires its value
is high, but each time it is used its strength decreases, as can be seen
in the graph in figure~\ref{curves}. Neurons which do not belong
to the winning neighbourhood give an input of $S(t)=0$ to the synapse.
This has the affect of causing the
efficacy of the synapse to increase, or `forget' some of
its inhibition, dishabituation.


\subsection{Putting it all together}

By attaching an habituable synapse to each of the neurons
in the SOM, a novelty filter is produced. The network is
shown in figure~\ref{hsom}. The only remaining question
is how the constants $\alpha$ and $\tau$ should be chosen.
In order for the network to learn quickly, the synapse of
the winning neuron should habituate rapidly. By choosing a
value of $\tau = 3.33$, the synapse decreases to below 90\%
of its original value within 5 iterations. The neighbourhood
neurons, which recognise similar perceptions, have a smaller
amount of habituation, $\tau = 14.33$ and the other neurons,
which are forgetting, have a longer time period, $\tau = 100$.
This is because we do not want the network to forget perceptions
too rapidly. Using this value a perception will recover from
complete habituation in about 280 presentations.

\section{\textbf{Experiments}}

The experiments presented
investigate the ability of the novelty filter to learn a model of an external environment
through periodic sonar scans taken while exploring, and to detect deviations from that model.
The effects of the forgetting mechanism are demonstrated.

\subsection{\textbf{The Robot}}

\begin{figure}
\centering
\includegraphics[width=.45\textwidth]{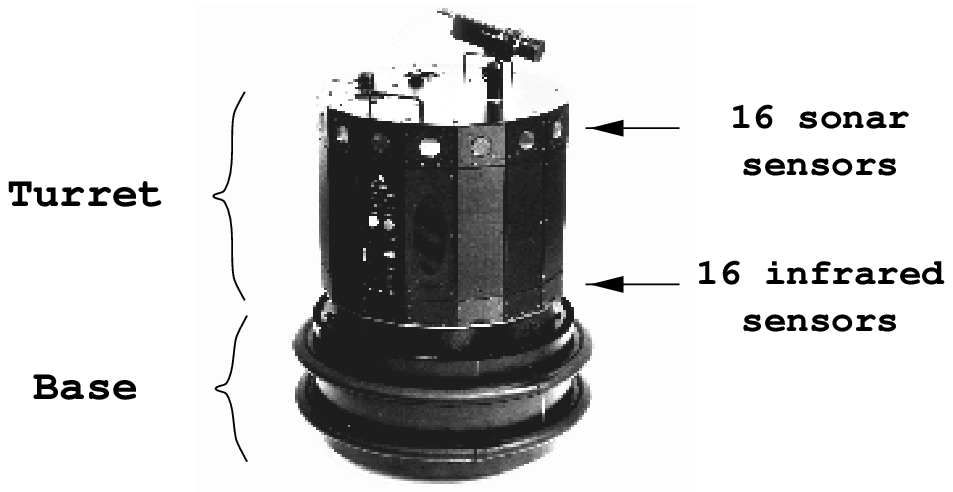}
\caption{ \textsf{ \small {The Nomad 200 mobile robot.}}}
\label{FortyTwo}
\end{figure}

A Nomad~200 mobile robot (shown in figure~\ref{FortyTwo}) was used to perform the experiments. The
band of infra-red sensors mounted at the bottom of the turret of the robot were 
used to perform
a pre--trained wall--following routine~\cite{Nehmzow94a}, and the 
16 sonar sensors at the top of the turret were used to provide perceptions
of the robot's environment. The angle between the turret and base 
of the robot was kept fixed. 
The input vector to the novelty filter
consisted of the 16 sonar sensors, each normalised to be between 0 and 1, 
were thresholded at about 4~metres.
The readings were inverted so that inputs from sonar responses received from closer objects were greater.

\subsection{\textbf{Experimental Procedure \label{Description}}}

\begin{figure*}
\centering
\includegraphics[width=\textwidth,height=.4\textheight]{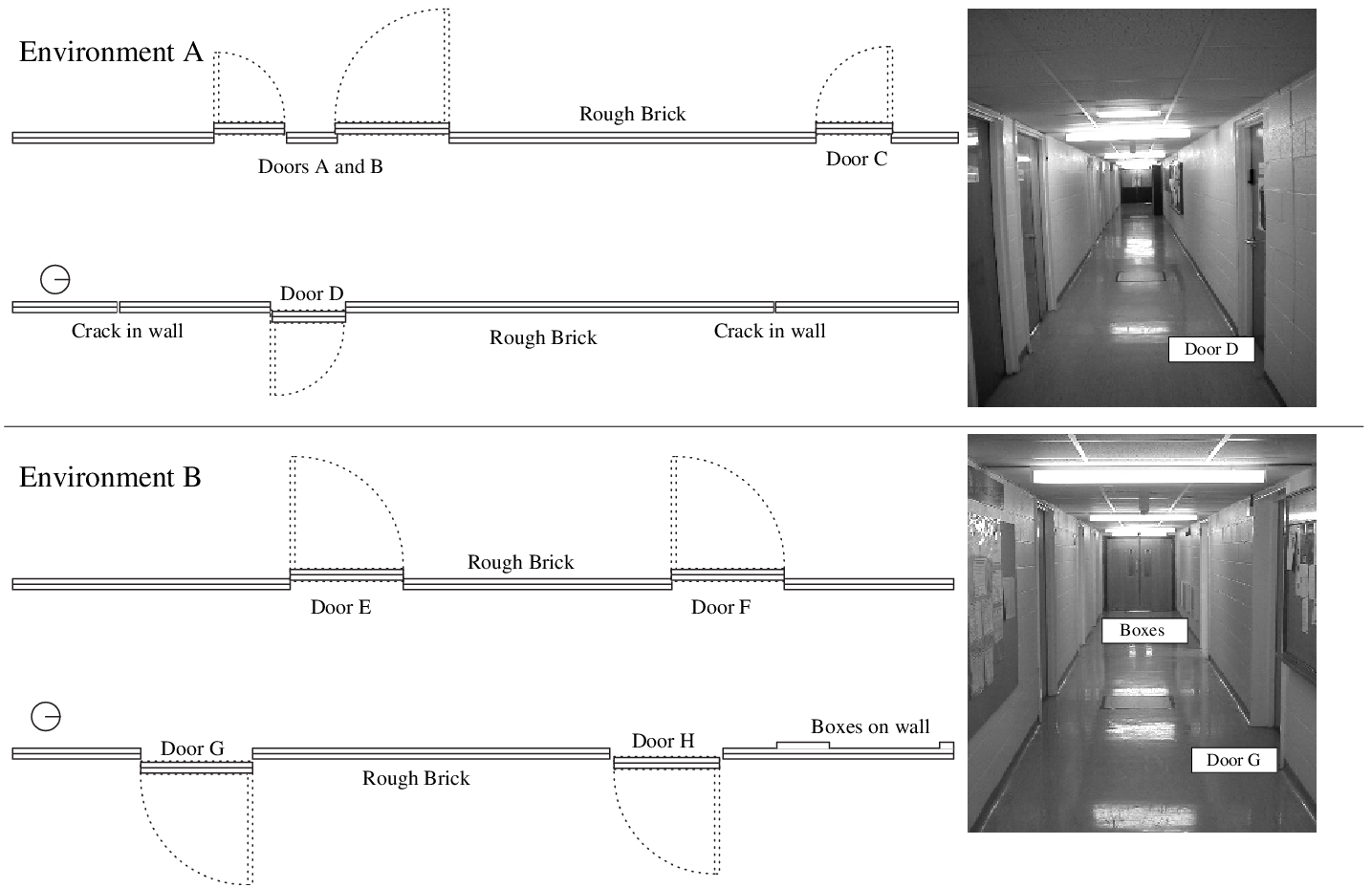}
\caption{ \textsf{ \small {Diagrams of the environments used.
The robot is shown facing in the direction of travel adjacent to the
wall that it followed. The environments are two similar sections
of corridor. 
The photographs show the environments as they appear from the
starting position of the robot. The notice boards which can be seen
in environment B are above the height of the robot's sonar sensors,
and were therefore not detected.}}}
\label{envs}
\end{figure*}

The experiments each consisted of a number of trials. In each trial,
the robot started from an arbitrarily chosen starting point, and moved using
a wall--following behaviour. Every 10\,cm along the route, the smoothed
readings from the sonar sensors were presented to the novelty filter, which
produced a novelty reading. Once the robot had travelled 10\,m it stopped
and saved the neural network weights. The robot was then returned to the
starting point using manual control and the same procedure repeated with
the updated network weights.

After each training trial, where the novelty filter learned about the
environment, the learning mechanism was turned off and a non-learning
trial performed. The sonar inputs still generated output from the novelty
filter to record the novelty of perceptions, but the robot did not learn.

\subsection{Environments}

Two environments were used in the experiments, together with a control 
environment for training. The two environments are shown in figure~\ref{envs}.
They are similar sections of corridor on the
second floor of the Computer Science building at the University
of Manchester. The corridors are 1.7\,m wide and have walls made from
painted breezeblock. Doors made of varnished wood lead from the corridors
into offices.

\section{\textbf{Results \label{Results}}}

\subsection {Experiment One \label{Res1}}

\begin{figure*}
\centering
\vspace{-15mm}
\includegraphics[height=.75\textwidth,width=.55\textheight,angle=90]{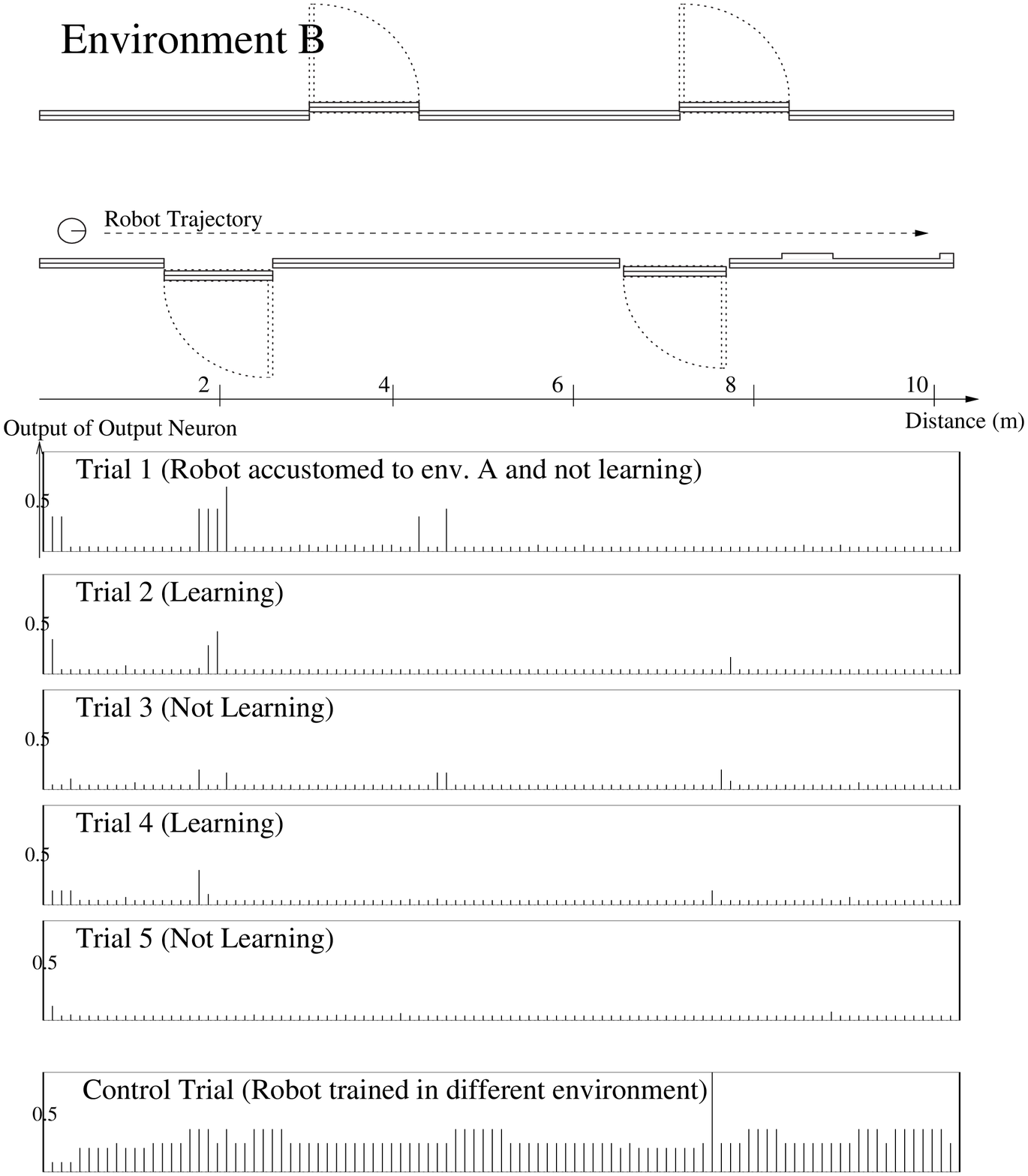} %
\includegraphics[height=.75\textwidth,width=.55\textheight,angle=90]{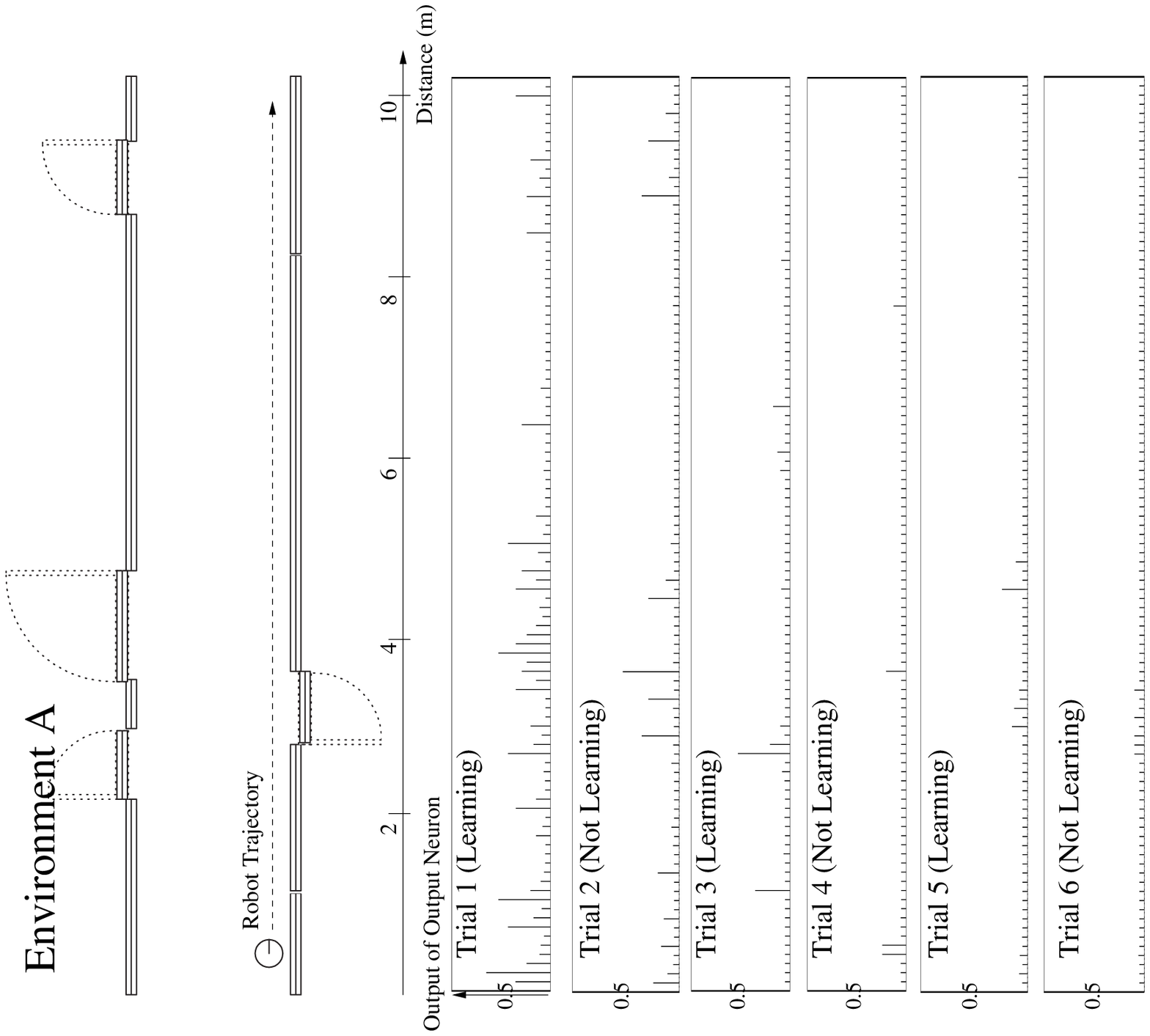} 
\vspace{-15mm}
\caption{ \textsf{ \small {The results of the first experiment. The graphs
on the left show the response of the output neuron of the novelty filter as the robot moves within
environment A when learning and not learning. Once it has stopped detecting
novelty features (so that the activity of the output neuron is low), the
robot was moved into environment B. The results of this are
shown on the right, and are discussed in section~\ref{Res1}. The final
picture on the right shows the results of investigating environment B
after prior training in a completely different control environment.}}}
\label{Exp1}
\end{figure*}

The first experiment aimed to demonstrate the ability of the novelty filter
to learn a representation of an environment and recognise novel features, so
that the robot could be used as an inspection agent. The novelty filter was
initialised randomly and then the robot was put into environment A. The left
of figure~\ref{Exp1} shows the results of this. The figures show the response
of the output neuron to the input vector of sonar readings that it receives
every 10\,cm along the route it travelled. At the top of the figure is a diagram of the
environment that the robot was travelling in at the time. Initially it can be seen 
that everything is novel, but where the robot perceives only wall, it rapidly
learns to recognise this. The next thing that it notices is the crack in the
wall on its right and then the doors. It is interesting to note that the robot 
finds the first crack more 
novel in the third run than in the second. This is because perceptions of such small features
vary greatly depending upon the precise position of the robot.
Only the cracks and the doorways are
highlighted in the later trials. After two more learning
trials, the novelty filter has learned an accurate representation of the
environment, as can be seen from the lack of response from the output neuron
in trial~6.

Once the novelty filter stopped finding anything novel in environment A, the
robot was moved into environment B. This is a similar environment to A (see figure~\ref{envs}).
The right of figure~\ref{Exp1} shows the results of this. The only things which
the robot finds to be novel in this environment are the perceptions of the
doorways. This is because the doors are inset further into the wall in this environment.
The control trial demonstrates the responses of the novelty filter when the
robot is put into environment B after training in a control environment.
For this the robot was driven around in an open area, travelling 
close to a wall, into the open space and back to the wall. It can be
seen that the robot finds the environment to be considerably more novel
after this training.

\subsection {Experiment Two \label{Res2}}

\begin{figure*}
\centering
\vspace{-15mm}
\includegraphics[height=.75\textwidth,width=.55\textheight,angle=90]{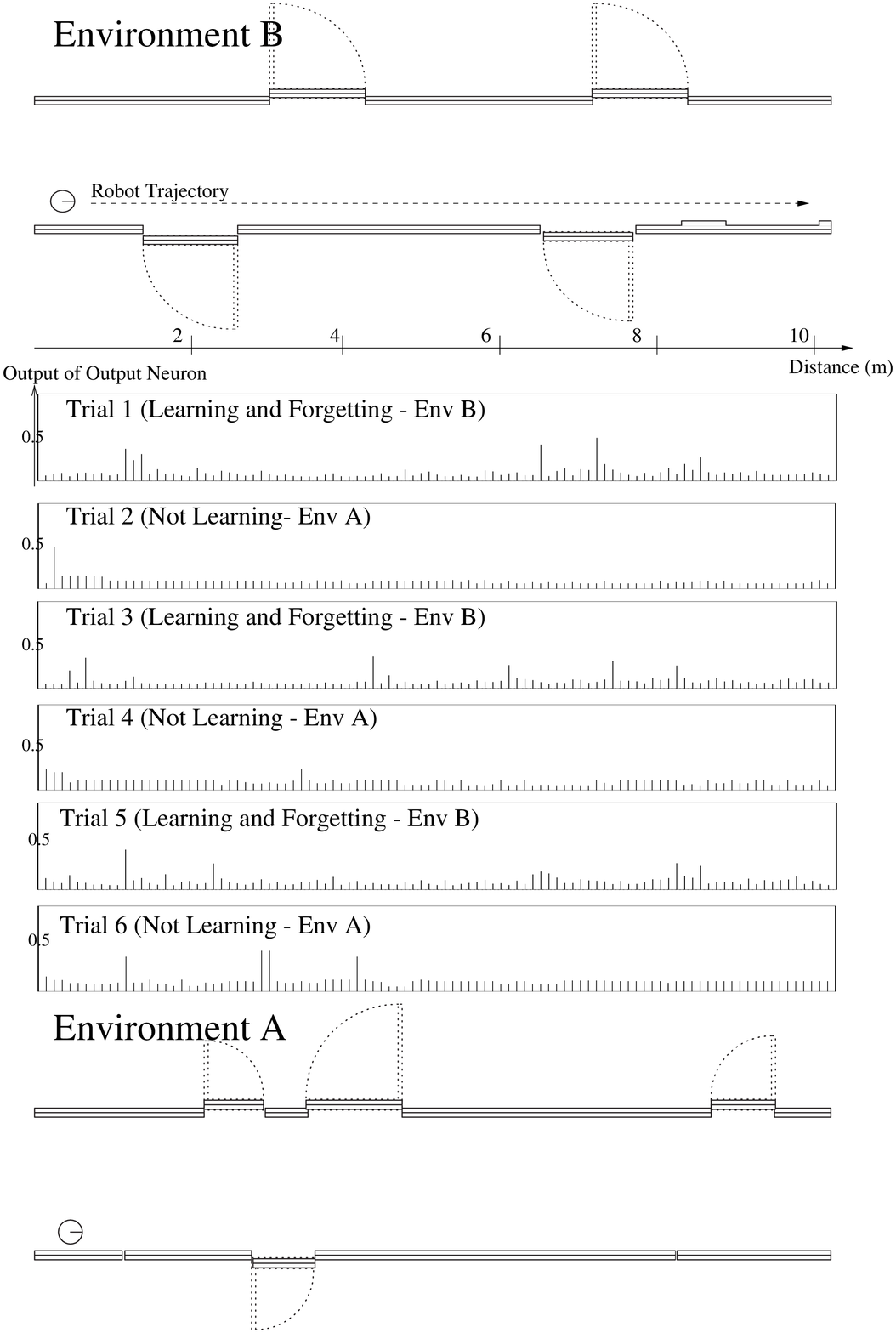} %
\includegraphics[height=.75\textwidth,width=.55\textheight,angle=90]{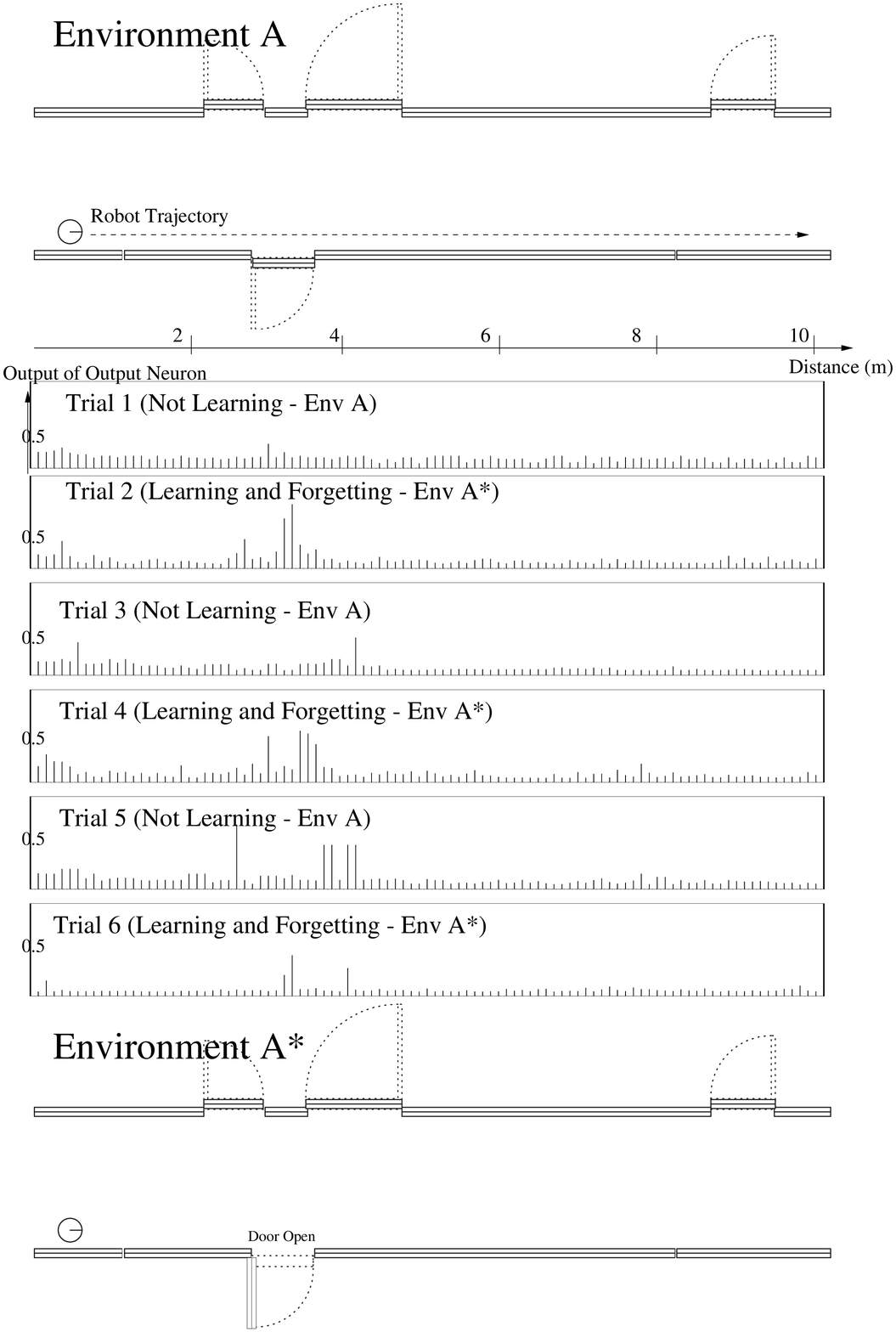} 
\vspace{-15mm}
\caption{ \textsf{ \small {A demonstration of the effects of forgetting. In the figure 
on the left the robot was accustomed to environment A. The environment was then
changed by opening a door (shown at the bottom of the figure) and the robot
learned this new environment, with forgetting turned on. It can be seen that after every 
exploration, the trial with the door closed finds more novelty in this feature.
A similar experiment is shown on the right, but using environments A and B. The
results are similar and are discussed in section~\ref{Res2}.}}}
\label{Exp2}
\end{figure*}

The second pair of experiments (shown in figure~\ref{Exp2}) were designed to 
show the behaviour of the forgetting
part of the novelty filter. Two different experiments were performed. In both 
the network weights learned when exploring environment A were used. The
first trial shows that the novelty filter had learned about this environment,
since the robot did not find anything novel.
A door in the environment was then opened (shown as Environment A* at the bottom 
of figure~\ref{Exp2}), and the robot learned about this new environment. A cardboard
box was placed in the doorway. This was of sufficient height to be seen by the
infra-red sensors which were responsible for the wall-following, but not by
the sonar sensors. After
each learning trial in this environment, the door was closed and a non-learning
trial in environment A was performed. The figures show that while no other features
are detected, the open door is initially novel, but is learned over the three
trials, and the closed door is initially recognised but is found progressively
more novel as the novelty filter forgets about this perception (since the filter
is not learning when the robot perceives the closed door).

The right of figure~\ref{Exp2} shows the second experiment. A similar technique 
was used, again starting with the network weights learned in environment A in the first experiment.
The robot learned about environment B, and after each learning trial was returned to 
environment A for a non-learning trial. Similar results can be seen - the robot
initially finds parts of environment B novel, but learns to recognise it over the trials,
while environment A, which is recognised at first, becomes more novel.
Obviously, only particularly features of environment A are found novel, those which
are not also seen in environment B. These
are the crack in the wall near the beginning of the environment and the doorway,
which is set into the wall less than those in environment B.

\section{\textbf{Summary and Conclusions}}

The experiments described have demonstrated that the novelty filter can
be used to learn a model of an environment and detect deviations from 
this model. The second experiment demonstrated the ability of the
filter to  forget perceptions that have been learned previously.
This means that the novelty filter will find novel features which are
seen only occasionally or not seen for a long time.
Therefore it can be trained in dynamic environments, where unforeseen
and undesirable perceptions, such as people walking past the robot, can
occur.

There are a number of areas which need further investigation. The integration
of a number of additional sensory systems will allow the filter to be more
widely applicable. In particular, the output of a monochrome CCD camera will
be used. The images will need to be extensively preprocessed before being
presented to the novelty filter to reduce computational time. In addition,
an investigation into alternatives to the Self-Organising Map (SOM) used in this
work is underway. There are a number of well documented problems with
the SOM, such as the fact that the size of the network needs to be 
pre-determined, which means that the network can fill up, so that
novel stimuli are incorrectly recognised as familiar. 
One possible solution is to use a growing network such as the 
Growing Neural Gas of Fritzke~\cite{Fritzke95}, another is to use
a Mixture of Experts~\cite{Jordan94}, with each expert learning a 
representation of a particular feature and voting on the novelty
of perceptions. A committee of networks, where networks of varying
sizes and training regimes vote on the response to a particular
input~\cite{Krogh95,Perrone93} could also be used.

\section*{\textbf{Acknowledgements}}
This research is supported by a UK EPSRC Studentship.

\bibliography{thebib,Manbib}
\bibliographystyle{plain}

\end{document}